\pdfoutput=1

\newif\ifarXiv
\arXivtrue

\ifarXiv
  \documentclass[a4paper]{amsart}

  \usepackage[T1]{fontenc}
  \usepackage{crimson}

  \usepackage{csquotes}

  \usepackage[%
    maxbibnames = 99,
    giveninits  = true,
    backend     = biber,
  ]{biblatex}
\else
  \documentclass{notices}
\fi

\usepackage         {amsfonts}
\usepackage         {amssymb}
\usepackage         {amsmath}
\usepackage[english]{babel}
\usepackage         {booktabs}
\usepackage         {dsfont}
\usepackage         {graphicx}
\usepackage         {microtype}
\usepackage         {mleftright}
\usepackage         {nicefrac}
\usepackage         {paralist}
\usepackage         {pgfplots}
\usepackage         {tabularx}
\usepackage         {tikz}
\usepackage         {tikz-3dplot}
\usepackage         {xspace}

\usepackage[subrefformat = parens]{subcaption}

\usetikzlibrary{arrows.meta}
\usetikzlibrary{chains}
\usetikzlibrary{positioning}

\definecolor{cardinal}{RGB}{196, 30, 58}
\definecolor{bleu}    {RGB}{ 49,140,231}

\pgfplotsset{
  compat = 1.18
}

\ifarXiv%
  \usepackage{hyperref}
  \hypersetup{
    colorlinks = true,
    linkcolor  = bleu,
    urlcolor   = bleu,
    citecolor  = bleu,
  }
\fi

\usepackage[noabbrev, capitalise, nameinlink]{cleveref}

\Crefname{equation}{Eq.}{Eqs.}

\newcommand{\indicator}{\ensuremath{\mathds{1}}\xspace}
\newcommand{\integers} {\ensuremath{\mathds{Z}}\xspace}
\newcommand{\reals}    {\ensuremath{\mathds{R}}\xspace}
\newcommand{\sphere}   {\ensuremath{\mathds{S}}\xspace}

\ifarXiv%
  \title[Topology meets Machine Learning]{%
    Topology meets Machine Learning:\\
    An Introduction using the Euler Characteristic Transform%
  }
\else%
  \title{%
    Topology meets Machine Learning: An Introduction using the Euler
    Characteristic Transform%
  }
\fi

\ifarXiv%
  \usepackage{amsaddr}
  \author{Bastian Rieck}
  \address{%
    Department of Computer Science, University of Fribourg, Switzerland
  }
  \email{\url{bastian.grossenbacher@unifr.ch}}
\else%
  \author{%
    Bastian Rieck%
    \affil{%
      Bastian Rieck is a professor of computer science at University of
      Fribourg. His e-mail address is \url{bastian.grossenbacher@unifr.ch}.
    }%
  }
\fi

\begin{document}

\maketitle

\section*{}

Machine learning is shaping up to be \emph{the} transformative
technology of our times:
Many of us have interacted with models like ChatGPT, new breakthroughs
in applications like healthcare are announced on an almost
daily basis, and novel avenues for integrating these tools into scientific
research are opening up, with some mathematicians already using large
language models as proof assistants~\cites{Ghrist25a, Wong24a}.

Written for an audience of mathematicians with limited prior
exposure to machine learning, this article aims to
dispel some myths about the field, hoping to make it more welcoming and
open.
Indeed, from the outside, machine learning might look like a homogeneous
entity, but in fact, the field is highly diverse, like mathematics
itself.
While the main thrust of the field arises from engineering advances,
with bigger and better models, there is also plenty of space
for mathematics and mathematicians.
As a running example of how mathematics---beyond linear algebra and
statistics, the classical drivers of machine learning---may enrich the
field, this article focuses on \emph{topology}, which recently started
providing novel insights into the foundations of machine learning:
\emph{Point-set topology}, harnessing concepts like neighborhoods, can
be used to extend existing algorithms from graphs to cell
complexes~\cite{Hajij20a}.
\emph{Algebraic topology}, making use of effective invariants like
homology, improves the results of models for 3D shape
reconstruction~\cite{Waibel22a}.
Finally, \emph{differential topology}, providing tools to study smooth
properties of data, results in efficient methods for analyzing
geometric simplicial complexes~\cite{Maggs24a}.
These~(and many more) research strands are finding a home in the nascent
field of \emph{topological deep learning}~\cite{Papamarkou24a}.

\section{What is Machine Learning?}

Before diving into concrete examples, let us first take a step back and
introduce the field of machine learning.
Machine learning is broadly construed as the art and science of
employing algorithms to solve tasks, but with the added restriction that
the algorithm should---in a certain sense that will become clear
below---adapt itself to the data at hand.
To substantiate this informal definition, we need to introduce some
terms, beginning with the notion of a \emph{feature}, which denotes a quantity
derived from ``raw'' input data. For instance, given a set of chemical
structures, the number of carbon atoms could be a feature.
Adding other features---such as the number of oxygen atoms, ring
structures, or aromatic bounds---then yields a high-dimensional
representation of a chemical structure in the form of a \emph{feature
vector}, which, for convenience, we assume to live in some $\reals^d$.

Machine-learning algorithms may thus, on a very high level, be seen as
functions of the form $f\colon\reals^d \to \mathcal{D}$, where
$\mathcal{D}$ indicates some domain of interest.
When $\mathcal{D} = \reals$, we say that this is a \emph{regression
task},\footnote{%
  This also generalizes to predicting more than one value.
}
whereas when $\mathcal{D}$ is a set, we are dealing with
a \emph{classification task}.
As a running example, suppose we are interested in classifying chemical
structures. The domain~$\mathcal{D}$ could
consist of the labels ``toxic'' and ``harmless.'' Since we have access
to labels, our classification task is an example of \emph{supervised machine learning}.\footnote{%
  In \emph{unsupervised machine learning}, by contrast, input data are
  not labeled, requiring models to ``learn'' characteristic properties
  or patterns of the data-generation process such as clusters.
}
Not every function~$f$ is equally useful in this context, though.
If~$f$ would always predict ``toxic,'' it would not be a suitable function
for our task~(even though its prediction might be the safest bet).
The allure of machine learning lies in the fact that it provides
a framework to \emph{learn} a suitable function~$f$ by presenting the
algorithm with numerous examples of different chemical structures, in the hope
that with sufficient data, the underlying mechanism(s) driving toxicity can
be derived. Thus, when we show this function a new example, its guess as
to whether it is toxic or not is based on all previously-seen examples.
The field of machine learning has developed many suitable models for
finding or approximating such functions.
Chief among those is the concept of \emph{deep neural networks}.\footnote{%
  See Schmidhuber~\cite{Schmidhuber22a} for a ``deep dive'' into the
  respective algorithms and their origin stories.
}
Deep-learning models started a veritable revolution in some fields like
computer vision, mostly because
\begin{inparaenum}[(i)]
  \item they obviate the need for ``hand-crafted'' features, as used
    at the beginning of this section, and
  \item they consist of standalone building blocks, i.e., \emph{layers}, that can
    be easily combined to build new models for solving domain-specific problems.
\end{inparaenum}

Initially, all layers of a deep neural network start with a random set
of parameters or \emph{weights}, which are then subsequently adjusted to
minimize a \emph{loss function}.
A simple deep neural network---a \emph{fully-connected neural
network}---produces an \emph{output}~$y_i^{(l)} \in \reals^{d_l}$ based on an
\emph{input}~$y_i^{(l-1)} \in \reals^{d_{l-1}}$ via the recursion
\begin{equation}
  y_i^{(l)} := \sigma\left(W^{(l)} y_i^{(l-1)} + b^{(l)}\right),
\end{equation}
where $l \in \{1, \dotsc, L\}$ ranges over the layers, $W^{(l)}$ denotes
a $d_l \times d_{l-1}$ matrix of weights, $b^{(l)} \in \reals$ is
a \emph{bias term}, and $\sigma$ refers to a nonlinear \emph{activation
function}~(like a sigmoid or $\tanh$).
Any original input $y_i^{(0)} \in \reals^d$ is thus transformed in a nonlinear fashion,\footnote{%
  In practice, layers can also perform different operations or
  transformations to the input data.
}
resulting in a final \emph{prediction} $y_i := y_i^{(L)} \in
\reals^{d_L}$, which is often normalized to $[0, 1]$ using
a \emph{softmax} function, i.e.,
\begin{equation}
  [y_i]_j \mapsto \frac{\exp\left([y_i]_j\right)}{\sum_{k=1}^{d_L}\exp\left([y_i]_k\right)},
\end{equation}
where $[y_i]_j$ denotes the $j$th component of $y_i$.
In our running example, \emph{binary cross-entropy}~(BCE) would then be
a suitable loss function. Mapping $\{$``harmless,'' ``toxic''$\}$ to
$\{0, 1\}$, the BCE loss for $n$ predictions is
\begin{equation}
  \frac{1}{n} \sum_{i = 1}^{n} \widehat{y_i} \log(y_i) + (1 - \widehat{y_i}) \log(1 - y_i),
  \label{eq:BCE loss function example}
\end{equation}
where $\widehat{y_i} \in \{0, 1\}$ represents the true label of a sample
and~$y_i \in [0, 1]$ refers to the softmax-normalized prediction of the
neural network, which we interpret as the probability of the sample
being in class~$1$, i.e., being toxic.
The parameters of the neural network are now subsequently adjusted using
\emph{gradient descent} or similar procedures, with the goal of
minimizing \cref{eq:BCE loss function example}. To this end, we
repeatedly perform the prediction for a set of predefined samples, the
so-called \emph{training data set}, which is presented to the neural
network in equal-sized \emph{batches}~(as opposed to using the full data
set as an input, which would often be prohibitive in terms of memory
requirements).
Once we are satisfied with the results, we evaluate the prediction on
the \emph{test data set}, which crucially, must be kept separate from
the training data set.

A common myth concerning deep learning is that it ``just'' performs
curve-fitting, arguably in a highly-elaborate way. While this article
cannot possibly counter all such claims, it should be pointed out
that some neural networks can approximate functions in certain
function spaces arbitrarily well. This property is also known as
\emph{universal function approximation} and implies that a certain class
of neural networks is dense~(usually with respect to the
compact convergence topology) in the function space.
For instance, feedforward neural networks are known to be able to
approximate any Borel-measurable function between finite-dimensional
spaces~\cite{Hornik89a}. Similar theorems exist for other architectures
and this property is often invoked when discussing the merits of deep
learning: A deep neural network with sufficient data may be 
the simplest way to approximate functions that do not permit analytical
solutions.

The conceptual simplicity and modularity of deep neural networks is not
without its downsides, though.
To perform well, deep neural networks
typically require enormous amounts of data as well as a humongous number
of parameters. Unless specific provisions are taken, deep neural
networks are not energy-efficient, and their data requirements still
pose serious obstacles in many domains. In addition, neural networks are
still black-box models with opaque outputs. Despite research in
\emph{interpretable machine learning} aiming to improve this state of
affairs, there are still few models whose outputs can be easily checked
by human operators, for instance. Even \emph{large language
models}~(LLMs), arguably one of the most impressive feats of the field,
suffer from shortcomings and are incapable of assessing their own
outputs. Like other deep neural networks, they also remain vulnerable to
\emph{adversarial attacks}, i.e., inputs that are generated to
provoke a bad response, often without the original user of the model
being aware of it~\cite{Chakraborty21a}.
Adversarial examples for vision models can appear
innocuous to a human observer but can cause incorrect predictions:
For example, traffic signs can be slightly remodeled using adhesive tape, making
a vision system unable to detect them.
This dispels the common myths that
\begin{inparaenum}[(i)]
  \item deep neural networks are always robust, and that
  \item deep-learning research is complete~(in the sense that, moving
    forward, there will not be any novel conceptual insights).
\end{inparaenum}
On the contrary---there are still fundamental challenges that
necessitate an analysis of the very foundations of deep learning. It is
here that topology and mathematics in general can provide a new perspective.

\section{Euler Characteristics Galore}

\begin{figure}[tbp]
  \centering%
  \includegraphics[width=\linewidth]{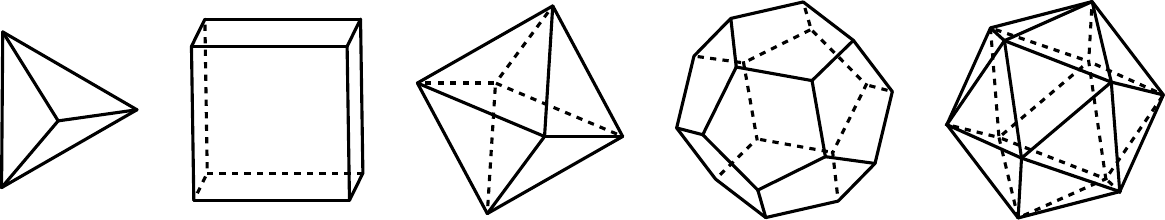}

  \medskip%
  \ifarXiv%
    \begin{tabular}{@{}lrrrr@{}}
  \else%
    \begin{tabularx}{\linewidth}{@{}Xrrrr@{}}
  \fi%
    \toprule
    \emph{Polyhedron} & $\chi$ & Vertices     & Edges      & Faces\\
    \midrule
    Tetrahedron       &     2 & 4             & 6          & 4    \\
    Cube              &     2 & 8             & 12         & 6    \\
    Octahedron        &     2 & 6             & 12         & 8    \\
    Dodecahedron      &     2 & 20            & 30         & 12   \\
    Icosahedron       &     2 & 12            & 30         & 20   \\
    \bottomrule
  \ifarXiv%
    \end{tabular}
  \else%
    \end{tabularx}
  \fi%
  \caption{%
    The five platonic solids~(tetrahedron, cube, octahedron,
    dodecahedron, and icosahedron) all have Euler characteristic~$\chi = 2$.
  }
  \label{fig:Polyhedra}
\end{figure}

\begin{figure*}[tbp]
  \centering
  \newsavebox{\enterprise}
  \savebox{\enterprise}{%
    \tdplotsetmaincoords{60}{135}%
    \begin{tikzpicture}[scale=3,tdplot_main_coords]%
      \draw[-stealth'] (0,0,0) -- (-1,0,0) node[above]{$x$};
      \draw[-stealth'] (0,0,0) -- ( 0,0,1) node[above]{$y$};
      \draw[-stealth'] (0,0,0) -- ( 0,1,0) node[right]{$z$};
      
      \node at (0,0) {\includegraphics[width=3.0cm]{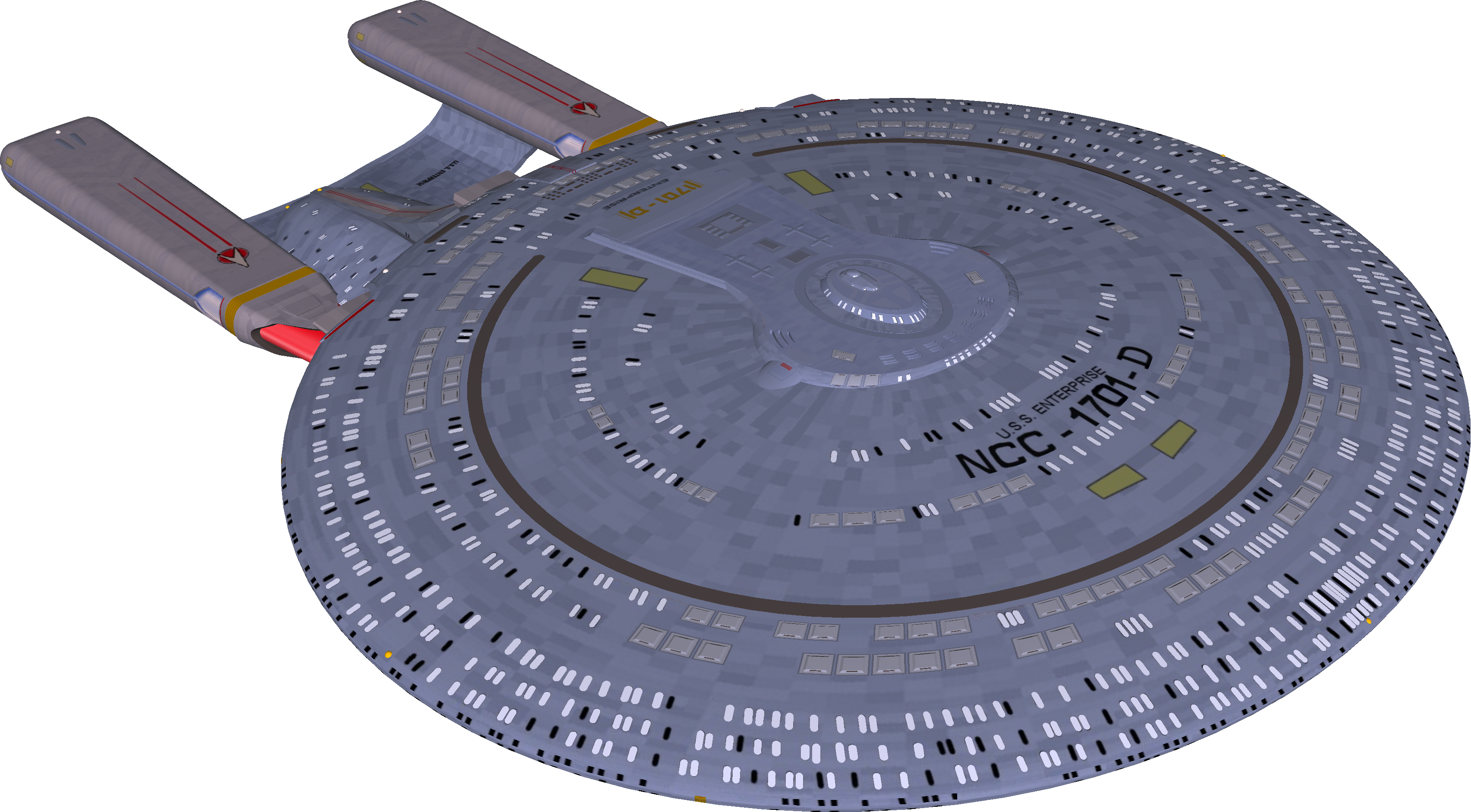}};%
    \end{tikzpicture}%
  }
  \ifarXiv%
  \resizebox{\linewidth}{!}{%
  \else%
  \fi
    \begin{tikzpicture}[start chain = main going right, node distance = 6.5mm]
      \tikzset{%
          block/.style = {%
            on chain,
            align          = center,
            inner sep      = 5.00pt,
            minimum height = 5.00cm,
            minimum width  = 2.00cm,
            draw           = black,
          },
          every label/.style    = {%
            font = \scriptsize,
          },
          >=stealth',
      }
      \node[block, label = below:{Geometric simplicial complex}] (Model) {%
        \usebox{\enterprise}%
      };

      \node[block, label = below:{Filtration in $y$-direction}] (Filtration) {%
        \includegraphics[width=4.0cm]{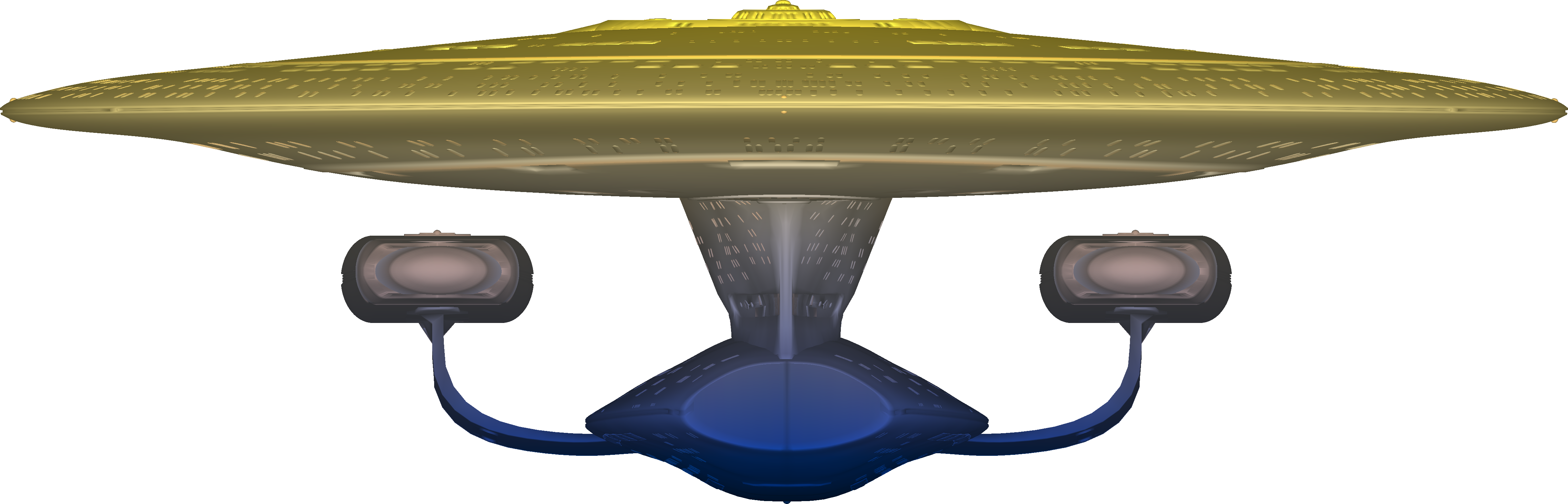}%
      };

      \node[block, label = below:{Euler Characteristic Curve}] (ECC) {%
        \includegraphics{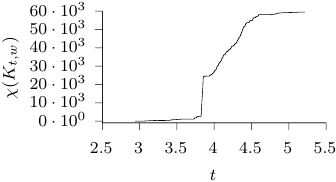}%
      };

      \draw[->] (Model)      edge (Filtration);
      \draw[->] (Filtration) edge (ECC);
    \end{tikzpicture}%
  \ifarXiv%
  }%
  \else%
  \fi
  \caption{%
    An illustration of the Euler Characteristic Curve calculation
    procedure. Starting from a 3D model, specified as a \emph{mesh} with
    vertices, edges, and faces, we perform a single filtration in the
    direction of the \mbox{$y$-axis}. The filtration values are shown in
    the cross-section of the shaded model, with warmer colors
    corresponding to larger filtration values. Finally, the Euler
    Characteristic Curve is plotted for all thresholds.\\[+0.125cm]
    \footnotesize%
    (The mesh depicts the ``Enterprise NCC-1701-D,'' from Star Trek: The
    Next Generation. It was originally created by Moreno Stefanuto and
    made available under a ``Free Standard'' license. This paper depicts
    a modified version by the author.)
  }
  \label{fig:ECC pipeline}
\end{figure*}

We focus our discussion on the \emph{Euler Characteristic
Transform}~(ECT). Being an invariant that bridges geometry and topology,
it is perfectly suited for such an overview article because it 
provides connection points for the largest number of researchers.
The ECT is based on the concept of the \emph{Euler characteristic}. This
integer-based quantity serves as a summary statistic of the ``shape'' of
a graph or simplicial complex.
We define an~(abstract) simplicial complex as a family of sets that is
closed under taking subsets~(also known as \emph{faces}).
A simplicial complex generalizes a graph by permitting more than mere
dyadic relations. We refer to the elements~(sets) of a simplicial
complex as its \emph{simplices} and, given a \emph{simplex}~$\{v_0,
\dots, v_i\}$, we say that its \emph{dimension} is~$i$.
A $p$-dimensional simplicial complex thus consists of simplices of
dimension up to and including~$p$.
For example, graphs can be considered $1$-dimensional
simplicial complexes, consisting of edges~(dimension~$1$) and
vertices~(dimension~$0$).
Given a \mbox{$p$-dimensional} simplicial complex~$K$, we write $K^{(i)}$ to
refer to its simplices of dimension~$i$. The \emph{Euler
characteristic} of~$K$ is then defined as an alternating sum of simplex
counts, i.e.,
\begin{equation}
  \chi(K) := \sum_{i=0}^{p} (-1)^i \left|K^{(i)}\right|.
\end{equation}
The Euler characteristic affords several equivalent definitions~(for instance,
in terms of the \emph{Betti numbers} of~$K$), generalizes to different
objects~(graphs, simplicial complexes, polyhedra, \dots), and is a \emph{topological
invariant}, meaning that if two spaces~(objects) are homotopy-equivalent, their
Euler characteristic is the same. This is depicted in \cref{fig:Polyhedra} by
means of the classical example of the five platonic solids. Since all
these spaces are homotopy-equivalent to the \mbox{$2$-sphere}, they all
share Euler characteristic~$2$.

While a single number~$\chi(K)$ is insufficient to fully characterize
a shape, it is possible to lift it to a multi-scale summary known as the
\emph{Euler Characteristic Transform}~(ECT), introduced by Turner et al.~\cite{Turner14a}.
The ECT requires a \emph{geometric simplicial complex}~$K$ in~$\reals^d$. We
can think of such a complex as having an associated coordinate~$x_v$ for
each vertex~$v$ of a simplex~$\sigma$ such that every face of~$\sigma$ is
in~$K$ and the intersection of two simplices in~$K$ is either
empty or a face of both.
Taking now any direction~$w \in \sphere^{d-1}$, i.e., any point on the
\mbox{$(d-1)$-sphere}, we obtain a real-valued function defined for all
simplices of~$K$ via 
\begin{equation}
  \begin{split}
    f_w\colon K &\to \reals\\
    \sigma & \mapsto \max_{v\in\sigma} \langle x_v, w \rangle,
  \end{split}
  \label{eq:Filtration}
\end{equation}
where $\langle \cdot, \cdot \rangle$ denotes the standard Euclidean
inner product.
In the parlance of computational topology, this function~$f_w$ can be used
to obtain a \emph{filtration} of~$K$ in terms of its subcomplexes: Given
a threshold~$t \in \reals$, we define $K_{t, w} := \{\sigma \in K \mid
f_w(\sigma) \leq t\}$.
Evaluating the Euler characteristic of each $K_{t,w}$ then yields the
\emph{Euler Characteristic Curve}~(ECC) associated to the direction~$w$.
Referring to the set of all finite geometric simplicial complexes as
$\mathcal{K}$, the ECC is an integer-valued function of the form
\begin{equation}
  \begin{split}
    \operatorname{ECC}\colon \mathcal{K} \times \sphere^{d-1} \times \reals &\to \integers\\
    (K, w, t)                                                               & \mapsto \chi(K_{t, w}).
  \end{split}
  \label{eq:ECC}
\end{equation}
The ECC thus maps a simplicial complex~$K$, together with a direction~$w$ and
a threshold~$t$, to the Euler characteristic of its corresponding subcomplex.
\Cref{fig:ECC pipeline} depicts the ECC calculation procedure for
a \emph{mesh}, i.e., a geometric simplicial complex in $\reals^3$, consisting
of $276,497$~vertices, $821,475$~edges, and $604,393$ faces~(triangles).
We consider a single filtration along the \mbox{$y$-axis} of the mesh, i.e., $w
= (0, 1, 0)$, sweeping the spaceship from bottom to top via $K_{t, w}$, where
$t \in [2.93, 5.22]$ as a consequence of the range of the vertex coordinates of
the mesh.
For each~$t$, we calculate the Euler characteristic~$\chi(K_{t, w})$,
leading to the ECC associated to the direction~$w$.

While more ``expressive'' than a single Euler characteristic, the ECC
still depends on the choice of direction. By using multiple directions,
we may hope to obtain a better representation of~$K$.
This leads to the \emph{Euler Characteristic Transform}, which is
defined as the function that assigns each direction~$w \in
\sphere^{d-1}$ to its corresponding ECC, i.e.,
\begin{equation}
  \begin{split}
    \operatorname{ECT}\colon \mathcal{K} \times \sphere^{d-1} & \to \integers^{\reals}\\
    (K,w)                                                     & \mapsto \operatorname{ECC}(K, w, \cdot),
  \end{split}
  \label{eq:ECT}
\end{equation}
where $\integers^{\reals}$ denotes the set of functions from~$\reals$
to~$\integers$.
Turner et al.~\cite{Turner14a} proved that the ECT serves as
a sufficient ``shape statistic,'' yielding an injective mapping for
dimensions~$2$ and $3$ if an infinite number of directions is used.
Thus, given two simplicial complexes $K \neq K'$, we have
$\operatorname{ECT}(K, \cdot) \neq \operatorname{ECT}(K', \cdot)$.
This result was later generalized to arbitrary dimensions by Ghrist et
al.~\cite{Ghrist18a} and Curry et al.~\cite{Curry22a},\footnote{%
  Despite their different publication dates, both works appeared concurrently
  as preprints.
}
who also showed that, somewhat surprisingly, a \emph{finite} number of
directions is sufficient to guarantee injectivity.

We briefly recapitulate the proof idea underlying Ghrist et
al.~\cite{Ghrist18a}, which draws upon Euler calculus on \emph{o-minimal
structures} to only permit inputs that are ``well-behaved'' or ``tame.''
An o-minimal structure $\mathcal{O} = \{\mathcal{O}_d\}$ over $\reals$ consists
of a collection of subsets $\mathcal{O}_d \subset \reals^d$, which are closed
under both intersection and complement. Moreover, $\mathcal{O}$ needs to
satisfy certain axioms, including\footnote{%
  This expository article strives for clarity. Interested readers are invited
  to dive into Curry et al.~\cite{Curry22a} and the references therein to learn
  more about o-minimal structures.
}
being closed under cross products and containing all algebraic sets. In
addition $\mathcal{O}_1$ must consist of finite unions of open intervals and
points.
The sets in $\mathcal{O}$ are called \emph{definable} or \emph{tame}.
Letting $X$ be a definable subset~(intuitively, $X$ can be seen as
a generalized variant of a geometric simplicial complex) of $\reals^d$, the \emph{constructible
functions} on $X$ are integer-valued functions with definable level
sets, denoted by $h\colon X \to \integers$.
Letting $\operatorname{CF}(X)$ refer to the set of constructible
functions on~$X$, the \emph{Euler integral} on~$X$ is the functional
$\int_X \cdot\, \mathrm{d}_{\chi}\colon \operatorname{CF}(X) \to
\integers$ that maps $\indicator_\sigma \mapsto (-1)^{\dim \sigma}$ for
each simplex~$\sigma$.
Given definable subsets $X, Y$ and a constructible function $k \in
\operatorname{CF}(X \times Y)$, the \emph{Radon transform} is defined as 
\begin{equation}
  \mathcal{R}_k h(y) := \int_X h(x) k(x, y) \mathrm{d}_\chi(x).
\end{equation}
Ghrist et al.~\cite{Ghrist18a} now show that the ECT can be considered
a Radon transform, using $X = \reals^d$ and $Y = \sphere^{d-1}
\times \reals$, with $k(x, (w, t))$ being the indicator function on
$\{(x,(w, t)) \mid \langle x, w \rangle \leq t \}$.
Since prior work already shows under which conditions one can recover
the input to a Radon transform, Ghrist et al.~\cite{Ghrist18a} obtain
a short, elegant proof of the fact that the ECT is not only injective
but also \emph{invertible}.

\section{Using the Euler Characteristic Transform in a Machine-Learning Model}

Given the invertibility results and the fact that the ECT is highly
efficient to implement, its integration into machine-learning models is
only logical.
Before we discuss how to use the ECT with a deep-learning model, let
us first ponder some of its practical uses with classical\footnote{%
  While this is the common nomenclature as used in the machine-learning
  community, it should be pointed out that some of these methods predate
  deep learning by a couple of years only. This indicates the rapid pace
  of the field.
}
machine-learning methods like \emph{support vector machines}. Central to
such algorithms is their use of hand-crafted features; while often
eschewed in deep learning,\footnote{%
  It is a common myth that classical machine-learning techniques
  are always outperformed by deep learning; in fact, there are many
  applications in which such techniques exhibit strong performance while
  remaining computationally efficient.
}
the power of such features should not be
underestimated---in particular when combined with methods like the ECT.

The results by Curry et al.~\cite{Curry22a} indicate that one could
use a finite set of directions to calculate the ECT. However, it is
unclear how to find such directions. This leaves us with the option to
\emph{sample} directions $W := \{w_1, \dots, w_k\} \subset
\sphere^{d-1}$, and consider the ECT restricted to $W$.
The hope is that the sample is sufficient to capture shape
characteristics. Another discretization applied in practice involves the
choice of thresholds. For example, if each coordinate $x_v$ has at most
unit norm, we know that $\langle x_v, w \rangle \in [-1, 1]$. Thus, we
may sample thresholds $T := \{t_1, \dots, t_l\} \subset [-1, 1]$ and
evaluate each ECC for a specific direction~$w$ only at thresholds
in~$T$.
This discretization allows us to represent the ECT as a matrix,
where columns are indexed by~$W$ and rows are indexed by~$T$.
In this representation, each column corresponds to the ECC for some
direction~$w$ and all ECCs are aligned in the sense that a row index~$i$
within any column corresponds to the \emph{same} threshold~$t_i \in [-1,
1]$.
An alternative discretization involves picking a set of~$l$
thresholds~$T_w$ for each direction~$w$ individually. While each column
of the resulting matrix still represents an ECC, the same row index~$i$
within columns now generally corresponds to \emph{different} thresholds,
depending on the set of directions~$W$.
Both discretization strategies have their merits. The first strategy is
preferable when the input data is contained in~(or normalized to) a unit
sphere; the second strategy can be useful in situations in which size
differences between individual ECCs do not matter.

\begin{figure}[tbp]
  \centering
  \subcaptionbox{Raw}{%
    \includegraphics[height=3.25cm]{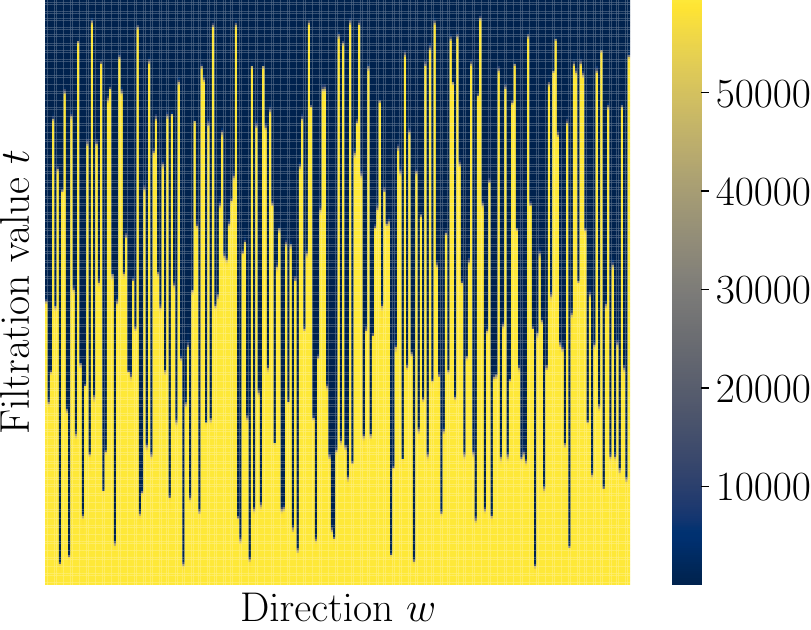}%
  }%
  \quad%
  \subcaptionbox{Column-normalized}{%
    \includegraphics[height=3.25cm]{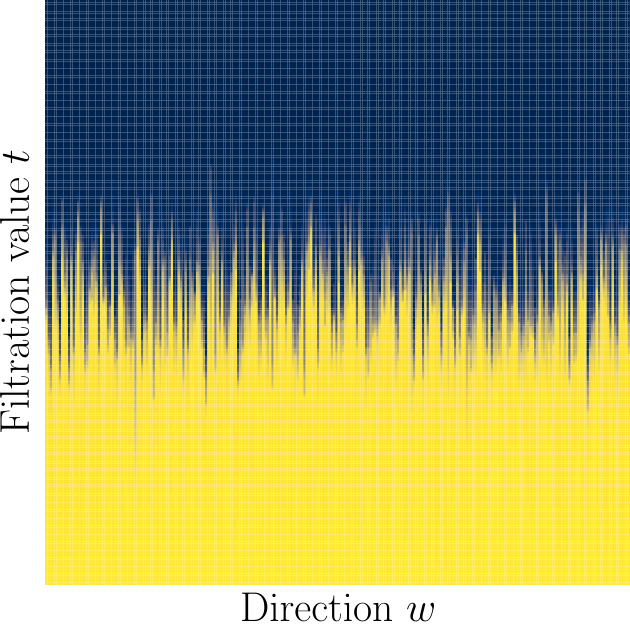}%
  }
  \caption{%
    Visualizing the ECT of the model in \cref{fig:ECC pipeline} as
    a discretized image using either the raw $t$-values~(left), or
    a re-scaled version~(right) where each column is normalized
    individually to $[-1, 1]$.
    Each $x$-axis shows $256$ randomly-selected directions, while each
    $y$-axis corresponds to the filtration values~$t$, with the lowest
    value starting at the top of the image. Every column in the image
    corresponds to an individual ECT, with the pixel color indicating
    the value for a specific filtration threshold~$t$.
  }
  \label{fig:ECT heatmaps}
\end{figure}

Regardless of the discretization strategy, the resulting matrix can be
viewed as an \emph{image}, although care must be taken to understand
that the ordering of directions, i.e., the columns of the matrix/image,
is entirely arbitrary. \Cref{fig:ECT heatmaps} depicts this visualization for
$256$ randomly-selected directions.  When using the ``raw'' values
of~$t$ to index the rows, we observe substantially more zeros. This is
because we chose thresholds based on the global minimum and maximum
across \emph{all} filtrations, meaning that any individual ECT, whose
range is typically much smaller than the global one, rises to its
respective maximum very quickly.
The column-normalized version, by contrast, exhibits more variability,
with different columns appearing to have a smaller width---this is
a perceptual illusion, though, arising from the fact that each column
now gradually increases in intensity.
Unlike the ``raw'' version, care must be taken when considering
differences between individual columns.

Setting aside the dissimilarities between the two variants, it is
possible to compare shapes by calculating an appropriate distance
between their respective ECT images. Such approaches work best when
shapes are aligned to a shared coordinate
system~\cite{Munch25a} since \emph{any} column permutation of the
resulting matrix describes the same ECT in the sense that
there is no canonical ordering of different directions in higher
dimensions.
The matrix representation can also be used directly as an input to
a machine-learning model by ``flattening'' the matrix, thus effectively
turning the matrix into a long feature vector. Similar to our initial
example with chemical structures, this process enables us to represent
a complex shape as a fixed-size vector. Such a representation crucially
depends on the choices of thresholds and directions, but is seen to
lead to good results in practice~\cite{Munch25a}.

How can we move beyond such hand-crafted features and use deep learning
to find task-specific ECTs?
The answer lies in picking a different representation, which enables us
to \emph{learn} an appropriate set of directions~$W$ as opposed to
\emph{sampling} it.
One obstacle to overcome here involves the fact that the ECT is a discrete
quantity based on step functions. Such functions are at odds with
deep-learning models, who prefer their inputs to be smooth and differentiable.
While different variants of the ECT exist, for instance a smooth one
obtained by mean-centering and integration~\cite{Munch25a}, we may also
apply another trick often found in machine-learning research: Instead of
working with the original definition of a function, we just work with
a smooth \emph{approximation} of it! While this technically solves a different
problem than what we originally set out to achieve, it often works
surprisingly well.
Regardless of the specific approximation method, this procedure will
permit us to use the ECT as a differentiable building
block of deep-learning models. The ECT may thus be said to
constitute an \emph{inductive bias}, a term that refers to the specific
assumptions built into a machine-learning model to affect its inner
workings and results.\footnote{%
  The quintessential example of an inductive bias involves the locality
  assumption of convolutional neural networks, i.e., the focus on small
  ``patches'' of an input image.
}

To approximate the ECT, we first notice that any ECC can
be written as a sum of \emph{indicator functions}, i.e., we rewrite
\cref{eq:ECC} as
\begin{equation}
  (K, w, t) \mapsto \sum_{i=0}^{p}(-1)^i\!\!\!\sum_{\sigma \in K_{t, w}^{(i)}}\!\!\!\!\indicator_{\leq t}( f_w(\sigma) ),
  \label{eq:ECC indicator functions}
\end{equation}
where $K_{t,w}^{(i)}$ denotes all \mbox{$i$-simplices} of $K_{t,w}$,
and~$f_w$ refers to the filtration from \cref{eq:Filtration}. Equivalently,
\cref{eq:ECT} permits a rephrasing via indicator functions.
So far, this is still an exact expression. The approximation comes into
play when we notice that an indicator function can be replaced by
a \emph{sigmoid function}, i.e., $S(x) = \nicefrac{1}{1 + \exp(-x)}$.
This lets us define an approximate ECC by rewriting \cref{eq:ECC
indicator functions} as
\begin{equation}
  (K, w, t) \mapsto \sum_{i=0}^{p}(-1)^i\!\!\!\sum_{\sigma \in K_{t, w}^{(i)}}\!\!\!\!S\mleft(\lambda\mleft(t - f_w(\sigma)\mright)\mright),
  \label{eq:ECC sigmoid functions}
\end{equation}
where $\lambda \in \reals$ denotes a scaling parameter that controls the
tightness of the approximation. \Cref{fig:Sigmoid approximation} depicts
this approximation for various values; we can see that $S(\lambda x)$ is
indeed a suitable~(smooth) replacement for an indicator function.
This minor modification has major advantages: \emph{Each} of the
summands in \cref{eq:ECC sigmoid functions} is differentiable with respect
to~$w$, $t$, and $f_w(\cdot)$. It is thus possible to use \cref{eq:ECC
sigmoid functions} to transform a geometrical simplicial complex in a way
that is fundamentally compatible with a deep-learning model.

Lest we get lost in implementation details, it is better to assume
a more elevated position and take stock of what we already have: Using
the approximation above, we have turned the ECT from a discrete function
into a function that \emph{continuously} depends on its input
parameters. In machine-learning terminology, we may thus treat this
calculation as a \emph{layer}. Assuming that all coordinates have at
most unit norm so that $\langle x_v, w\rangle \in [-1, 1]$, our ECT
layer has two hyperparameters, namely
\begin{inparaenum}[(i)]
  \item the number~$l$ of thresholds to discretize $[-1, 1]$, and
  \item the number~$k$ of directions.
\end{inparaenum}
The input to our layer is a geometric simplicial complex~$K$, and the
output is an \mbox{$l \times k$} matrix~(equivalently, we may apply such
a construction to all types of spaces that afford an Euler
characteristic, including \emph{cell complexes}, for instance).
Notice that when learning an ECT for a data set of different objects,
one matrix~(i.e., the discretized image of an ECT) for each sample in
the input batch is returned, resulting in a multidimensional array
output, which is also referred to as a \emph{tensor}.\footnote{%
  This terminology can be confusing for mathematicians at first since
  machine learning does not~(always) make use of any of the properties
  that would make up a ``mathematical'' tensor.
}
Given the continuous dependence on its input parameters, we refer
to the construction above as the \emph{Differentiable Euler
Characteristic Transform}~\cite{Roell24a}.

\begin{figure}[tbp]
  \centering
  \begin{tikzpicture}
    \begin{axis}[%
      axis x line        = bottom,
      axis y line        = left,
      enlarge y limits   = true,
      domain             = {-5:5},
      xmin               = -5,
      xmax               =  5,
      samples            = 200,
      width              = 0.50\linewidth,
      tick align         = outside,
    ]
      \addplot[black] {
        1 / (1 + exp(-x))
      };

      \addplot[black] {
        1 / (1 + exp(-2 * x))
      };

      \addplot[black] {
        1 / (1 + exp(-3 * x))
      };

      \addplot[black] {
        1 / (1 + exp(-4 * x))
      };

      \addplot[black] {
        1 / (1 + exp(-5 * x))
      };

      \addplot[domain =  0:5, dashed, very thick, cardinal] {1};
      \addplot[domain = -5:0, dashed, very thick, cardinal] {0};

    \end{axis}
  \end{tikzpicture}
  \caption{%
    An indicator function~(dashed) and its approximation using
    differently-scaled sigmoid functions, with $\lambda \in \{1, 2, 3, 4, 5\}$.
  }
  \label{fig:Sigmoid approximation}
\end{figure}

\section{Two Applications of Differentiable Euler Characteristic Transforms}

\begin{figure}[tbp]
  \centering
  \subcaptionbox{Original\label{sfig:ECT original}}{%
    \includegraphics[width=0.30\linewidth]{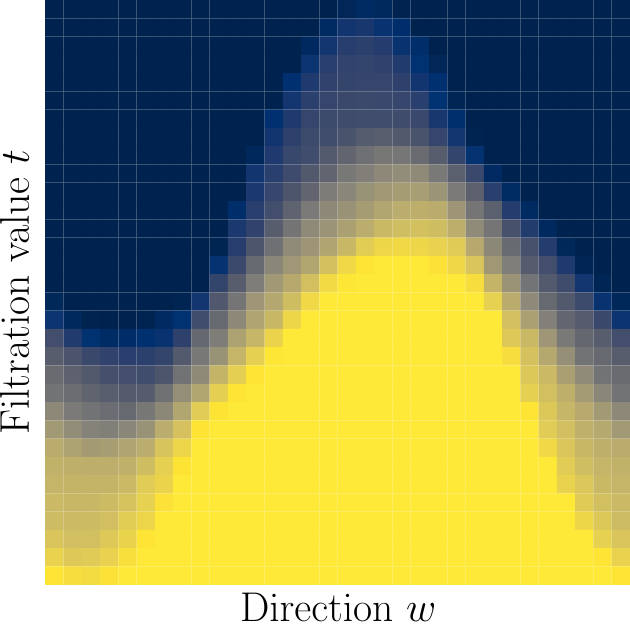}%
  }\quad%
  \subcaptionbox{Untrained\label{sfig:ECT untrained}}{%
    \includegraphics[width=0.30\linewidth]{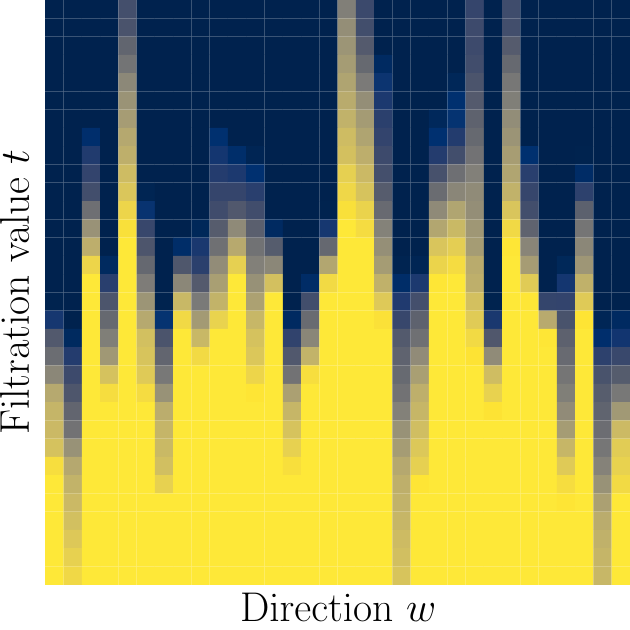}%
  }\quad%
  \subcaptionbox{Trained\label{sfig:ECT trained}}{%
    \includegraphics[width=0.30\linewidth]{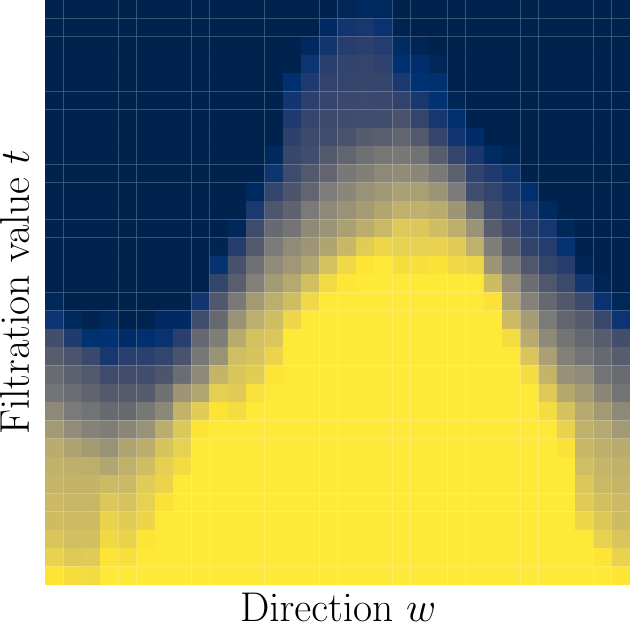}%
  }
  \caption{%
    \subref{sfig:ECT original} We can learn the directions required to
    match the ECT of a simple data set. \subref{sfig:ECT untrained}
    Starting with a set of random directions, our initial ``guess'' of
    the ECT looks nothing like the original. \subref{sfig:ECT trained}
    After several optimization steps, which minimize the dissimilarity
    between ``our'' ECT and the original one, we obtain a good
    approximation. We have thus learned a set of suitable directions for
    calculating the ECT.
  }
  \label{fig:Learning directions}
\end{figure}

The ECT layer we defined above gives rise to a fixed-size vectorial
representation, which may either be integrated into a larger neural
network---thus effectively handing off the ECT results for deeper
processing---or which can be used directly on its own. As an example of
the latter case, we can for instance \emph{compare} two outputs of an
ECT layer using a loss function. Treating the outputs as
high-dimensional vectors~$x$ and $y$, we can use the squared $l^2$-norm
of their difference, normalized by the dimension, as a criterion of how
well they are aligned. This
loss function is also known as the \emph{mean squared error}~(MSE).
It is commonly used to solve problems in \emph{representation
learning}, so it is perfectly suited for a small experiment: Suppose we
have a simple data set in~$\reals^2$ whose~(discretized) ECT we know.
Starting from a randomly-initialized set of directions, can we
\emph{learn} the ``best'' set of directions to align the two ECTs? It turns
out that the approximation scheme defined in \cref{eq:ECC sigmoid
functions} indeed permits solving such problems. In our toy
example, we start with $32$ uniformly-sampled directions, which we
can parametrize using a single angle.
We then sample the same number of angles from a normal distribution and
minimize the MSE loss between the ECT based on the random directions to
the ``original'' ECT. \cref{fig:Learning directions} depicts this
experiment, and we observe that only a couple of optimization
steps---using \emph{gradient descent}, for example---are required to
obtain a suitable approximation to our input ECT. Moreover, this
approximation gets better with longer training times and more
discretization steps.

\begin{figure}[tbp]
  \centering
  \pgfplotsset{
    point cloud/.style = {
      axis lines         = none,
      unit vector ratio* = 1 1 1,
    },
    width = 0.50\linewidth,
  }
  \tikzset{
    every mark/.append style = {
      mark size = 1pt,
    },
  }
  \subcaptionbox{Original\label{sfig:Point cloud original}}{%
    \begin{tikzpicture}
      \begin{axis}[point cloud]
        \addplot[only marks] table {Data/Double_annulus.txt};
      \end{axis}
    \end{tikzpicture}
  }\quad%
  \subcaptionbox{Untrained\label{sfig:Point cloud untrained}}{%
    \begin{tikzpicture}
      \begin{axis}[point cloud]
        \addplot[only marks] table {Data/Initial_point_cloud.txt};
      \end{axis}
    \end{tikzpicture}
  }\quad%
  \subcaptionbox{Trained\label{sfig:Point cloud trained}}{%
    \begin{tikzpicture}
      \begin{axis}[point cloud]
        \addplot[          only marks] table {Data/Double_annulus.txt};
        \addplot[cardinal, only marks] table {Data/Learned_point_cloud.txt};
      \end{axis}
    \end{tikzpicture}
  }
  \caption{%
    We can also learn the \emph{coordinates} of a point cloud based on
    the ECT. \subref{sfig:Point cloud original} We sample $100$ points
    from a double annulus. \subref{sfig:Point cloud untrained} The
    initial configuration of our coordinates is a random sample from
    a noisy circle. \subref{sfig:Point cloud trained} After minimizing
    the coordinates based on measuring the dissimilarity between the
    ``desired'' ECT and the current one, we are able to approximate the
    input point cloud closely~(\textcolor{cardinal}{red} points
    show the learned coordinates, black points the original
    ones).
  }
  \label{fig:Learning coordinates}
\end{figure}

However, learning the directions could be considered somewhat
solipsistic in that we derived a new representation~(the ECT) and then
showed that we can learn its parameters. Indeed, this experiment is more
of a ``smoke test,'' since learning the parameters of a representation
is the very basis of machine learning! To build a stronger experiment
showcasing the power of the ECT, let us consider learning
\emph{coordinates} instead. Recall that our approximation from
\cref{fig:Sigmoid approximation} permits us to also affect the
coordinates themselves: Any loss function that we evaluate will depend
on the directions \emph{and} the coordinates; thus, if we make the
coordinates a parameter of our layer, we can modify them similarly to
the example shown above. Using an MSE loss, this time we search for
input coordinates that minimize the differences of our ECT to a target
ECT~(machine-learning researchers also like to use the term \emph{ground
truth}). \cref{fig:Learning coordinates} depicts an example, using~$k = 256$
directions and $l = 256$. We again manage to learn suitable
coordinates that align our two point clouds. This can be very helpful
when working with coarser representations of data, i.e., we may use this
procedure to find the best approximation of a downsampled point cloud to
its original version, leading to an ECT-based compression algorithm. 

These tasks provide a glimpse of the versatility of the ECT. While
learning directions or coordinates is not necessarily a machine-learning
task \emph{sui generis}, a computational layer based on the ECT permits
more flexible applications. In conjunction with a loss term, we can employ
the ECT in a variety of tasks. For instance, we can use it to classify
\emph{geometric graphs}, i.e., low-dimensional simplicial complexes
whose vertices and edges are embedded in some~$\reals^n$. This task is typically
dominated by graph neural networks, which employ a mechanism called
\emph{message passing} that amounts to locally transporting information
via the edges of a graph.\footnote{%
  An expository article like this cannot possibly do justice to the
  large body of research available under that moniker. The reader is
  therefore invited to consult a recent position paper for more
  details~\cite{Velickovic23a}. 
}
The ECT offers a new and surprisingly
competitive method for classifying such graphs~\cite{Roell24a}.
Together with its
extremely small memory footprint---recall that the ECT is essentially
``just'' counting the constituent parts of an input object---this makes
the ECT an interesting paradigm to consider for new machine-learning
applications. Some of these applications are already discussed by Turner
et al.~\cite{Turner14a}, while others, including the experiments for
learning directions and coordinates, have been introduced in our recent
work~\cite{Roell24a}. As always, there is more work to be
done, some enticing directions being 
\begin{inparaenum}[(i)]
  \item an assessment of the theoretical expressivity of the ECT when it
    comes to distinguishing between geometric graphs, 
  \item the analysis of the \emph{inverse problem}, i.e., the
    reconstruction or generation of objects based on an ECT~(in the discrete setting),
    and
  \item algorithmic aspects that enable the ECT to perform efficiently
    in the context of high-volume geometry-based streaming data arising
    from LiDAR sensors, for instance.
\end{inparaenum}

\section{The Future of Topology in Machine Learning}

The ECT served as the \emph{leitmotif} of this article, demonstrating
how to connect concepts from applied topology and modern
machine-learning methods with relatively few hitches. The reader is
strongly recommended to check out an excellent overview article by
Munch~\cite{Munch25a} to learn more about it.
However, towards the end, let us briefly zoom out and consider the
larger picture~(knowing full well that predictions are hard, especially
when concerning the future). By design, this article could merely
scratch the surface, but the author firmly believes that topology and
topological concepts have a strong role to play in machine learning.
Many interesting directions are bound to fall into one of the following
three areas:
\begin{compactenum}
  \item Learning functions on topological spaces such as simplicial
    complexes or cell complexes.
  \item Building hybrid models that imbue neural networks with knowledge
    about topological structures in the data.
  \item Analyzing qualitative properties of neural networks.
\end{compactenum}
In the first area, topology can be used to generalize existing
machine-learning paradigms to a larger variety of input data sets; this
is appealing because not every data set ``lives'' in a nice Euclidean
space or a graph, and the inclusion of higher-order neighborhoods would
provide a shift in perspective, aiding knowledge discovery.
Researchers in \emph{topological data analysis}~(TDA) might feel
particularly comfortable in the second area since it is one of the
declared aims of TDA to understand topological features in data.
Nevertheless, hybrid models do not necessarily have to draw upon
concepts from TDA; new and daring methods could for instance focus on
computational aspects of Riemannian geometry like curvature or make use
of new invariants like metric-space magnitude.
Finally, the third area shifts the perspective and lets topology return
to its roots in that it can serve as a lens through which to study, for
instance, the training behavior of neural networks.
Understanding these training dynamics could lead to smaller, more
efficient models, but also shed some light on the soft underbelly of
deep-learning models, namely their susceptibility to unstable training
regimens, ``adversarial'' input data, or their propensity for
hallucinations.

Each of the three areas for new research has something enticing to
offer, not only for machine learning but also for~(applied) topology.
There is vast potential for new topology-aware models to
serve as proof assistants or even try to search for new conjectures and
counterexamples.
It is said that only those with their feet on rock can build castles in
the air. Topology can provide this rock upon which robust
machine-learning research can be built.
Thinking \emph{beyond} topology, the author hopes that this article may
also stir up the curiosity of mathematicians coming from other domains.
Machine-learning research will only benefit from more inquisitive minds and
there are countless things waiting to be discovered.
To get a taste of potential directions, readers are invited to peruse an
excellent expository article on deep learning~\cite{Higham19a}. The
author of this article also prepared additional scripts and literature
on the ECT, which are made available under
\url{https://topology.rocks/ect}.

\section*{Acknowledgments}

The author is indebted to Ernst Röell, Emily Simons, and the anonymous
referees for their helpful comments, which served to substantially
improve this article.
This work has received funding from the Swiss State Secretariat for
Education, Research, and Innovation~(SERI).

\ifarXiv%
  \clearpage%
  \printbibliography%
\else%
  \bibliography{main}
\fi%

\end{document}